\documentclass[letterpaper]{article} 
\usepackage{aaai23}  
\usepackage{times}  
\usepackage{helvet}  
\usepackage{courier}  
\usepackage[hyphens]{url}  
\usepackage{graphicx} 
\usepackage{natbib}  
\usepackage{caption} 
\usepackage[]{algpseudocode}
\usepackage{algorithmicx,algorithm}
\usepackage{amssymb}
\usepackage{amsmath}
\usepackage{booktabs}
\usepackage{multirow}
\DeclareMathOperator{\emb}{e}
\DeclareMathOperator{\unk}{unk}

\usepackage{newfloat}
\usepackage{listings}
\DeclareCaptionStyle{ruled}{labelfont=normalfont,labelsep=colon,strut=off} 
\floatstyle{ruled}
\newfloat{listing}{tb}{lst}{}
\floatname{listing}{Listing}
\title{Adversarial Word Dilution as Text Data Augmentation in Low-Resource Regime}
\author{
    Junfan Chen\textsuperscript{\rm 1},
    Richong Zhang\textsuperscript{\rm 1,2}\thanks{Corresponding author: zhangrc@act.buaa.edu.cn},
    Zheyan Luo\textsuperscript{\rm 1},
    Chunming Hu\textsuperscript{\rm 1,2},
    Yongyi Mao\textsuperscript{\rm 3}
}
\affiliations{
    \textsuperscript{\rm 1}SKLSDE, School of Computer Science and Engineering, Beihang University, Beijing, China \\
    \textsuperscript{\rm 2}Zhongguancun Laboratory, Beijing, China\\
    \textsuperscript{\rm 3}School of Electrical Engineering and Computer Science, University of Ottawa, Ottawa, Canada \\
    chenjf@act.buaa.edu.cn, zhangrc@act.buaa.edu.cn, akamya@buaa.edu.cn, hucm@buaa.edu.cn, ymao@uottawa.ca
}

\usepackage{bibentry}

\begin{document}

\maketitle

\begin{abstract}
Data augmentation is widely used in text classification, especially in the low-resource regime where a few examples for each class are available during training. Despite the success, generating data augmentations as hard positive examples that may increase their effectiveness is under-explored. This paper proposes an Adversarial Word Dilution (AWD) method that can generate hard positive examples as text data augmentations to train the low-resource text classification model efficiently. Our idea of augmenting the text data is to dilute the embedding of strong positive words by weighted mixing with unknown-word embedding, making the augmented inputs hard to be recognized as positive by the classification model. We adversarially learn the dilution weights through a constrained min-max optimization process with the guidance of the labels. Empirical studies on three benchmark datasets show that AWD can generate more effective data augmentations and outperform the state-of-the-art text data augmentation methods. The additional analysis demonstrates that the data augmentations generated by AWD are interpretable and can flexibly extend to new examples without further training. 
\end{abstract}
\section{Introduction}
Training effective text classification models often relies on sufficient precisely labeled data. However, it is common in some real-world scenarios that collecting plenty of valid text data is difficult, e.g., requiring considerable effort from human annotators. Enabling the model to learn efficiently with limited resources therefore becomes a practical need and hotspot in the industry and research communities. For example, text classification in few-shot learning that trains on small tasks composed of a few examples and tests on new tasks of unseen classes~\cite{Yu:18, Geng:19, ChenZM:22}, semi-supervised learning that provides a few labeled texts and plenty of unlabeled texts for training~\cite{MiyatoDG:17, XieDHL:20, LeeKH:21} and the low-resource regime that also provides a few labeled texts in training but no unlabeled texts available~\cite{WeiZ:19, GLZH:22}. Data augmentation is widely used in these tasks to increase data size and boost training and is often evaluated in the low-resource regime, which we will focus on in this paper.
\begin{figure}[ht]
	\begin{center}
		\begin{tabular}{c}
                \hspace{-.29cm}
		    \includegraphics[width=0.47\textwidth]{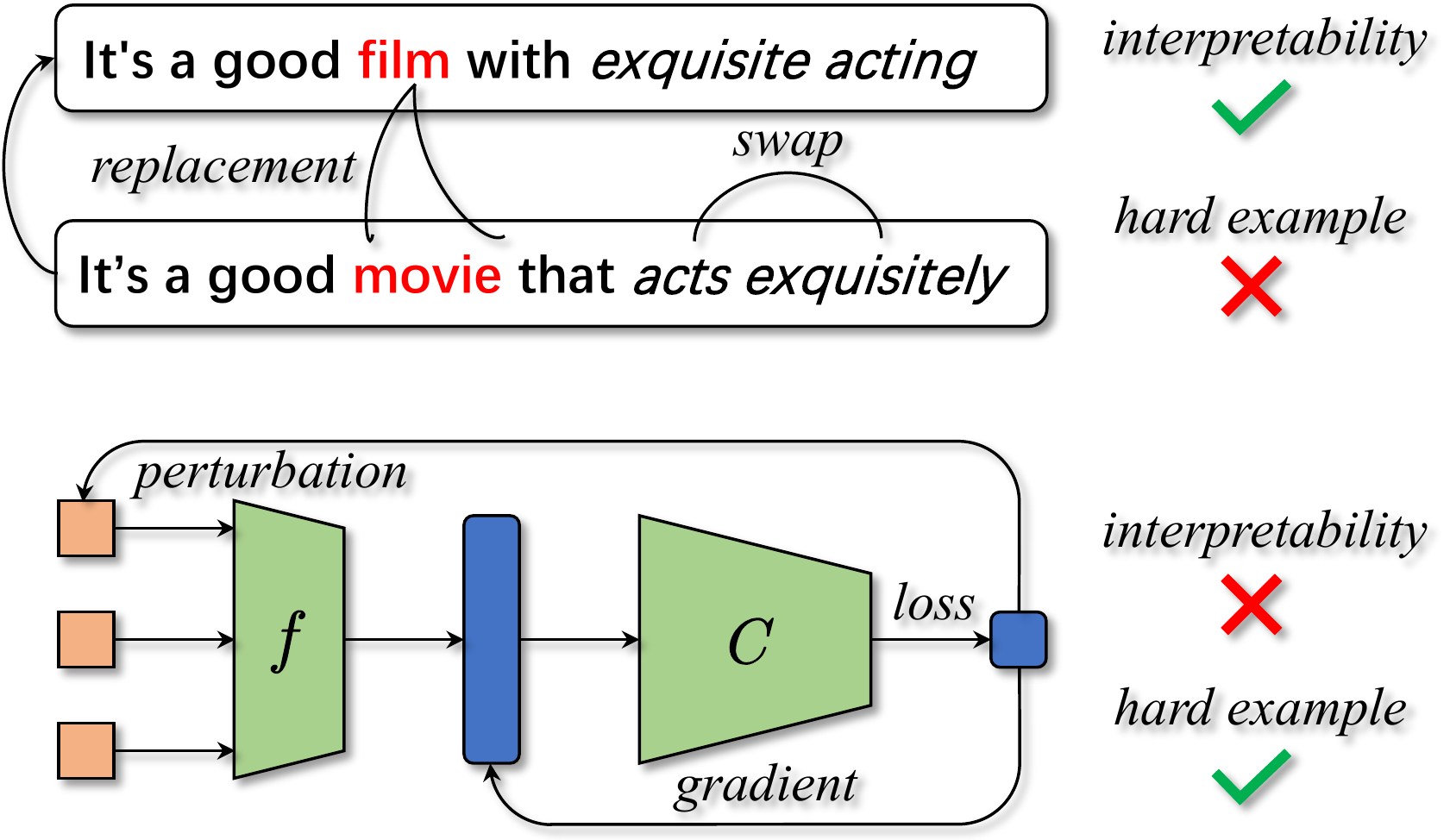}
		\end{tabular}
	\end{center}
	\caption{Illustration of contextual augmentation and representational augmentation (taking adversarial data augmentation as an example). The upper and lower part of the figure respectively show an example of contextual augmentation and the process of adversarial data augmentation. Contextual augmentation is interpretable but may not generate hard positive examples. Adversarial data augmentation can generate hard positive examples but is short of interpretability.}
	\label{fig:intro}
\end{figure}

Text data augmentation is challenged by the discrete nature and complex semantics of texts. Despite the challenges, two branches of textual data augmentation techniques have been explored and demonstrated effective: {\em contextual augmentation} and {\em representational augmentation}. The contextual augmentation methods augment the texts by replacing, inserting, deleting or swapping words~\cite{KolomiyetsBM:11, WangY:15, ArtetxeLAC:18, GaoZWXQCZL:19, WeiZ:19, MiaoLWT:20}, or paraphrasing the original texts~\cite{EdunovOAG:18, ChenTWG:19, KumarBBT:19, Thomas:21}. For example, in the upper part of Figure \ref{fig:intro}, given the text ``{\em It's a good movie that acts exquisitely},'' one of its contextual augmentations may be ``{\em It's a good film with exquisite acting}.'' The contextual augmentations are semantically interpretable because the modifications are expected to keep semantic consistency with the original texts. However, such heuristic modifications may produce simple data augmentations that bring little improvements or even degraded performance~\cite{ZhangZL:15}.

The representational augmentation methods generate augmented inputs by interpolation or perturbation of the word embedding or text representations~\cite{MiyatoDG:17, HsuTG:18, WuWQ0:19, ChenRLL:20, JiangHCLGZ:20, ZhuCGSGL:20, ChenJ:21}. One most intensively used perturbation method is adversarial data augmentation, which creates augmented examples by adding adversarial noise to the original representations~\cite{MiyatoDG:17, JiangHCLGZ:20, ZhuCGSGL:20}. As shown in the lower part of Figure \ref{fig:intro}, since the adversarial noise is generated by gradients minimizing the model objective, the adversarial data augmentations could be regarded as hard positive examples. However, the augmentations lack semantic interpretability because the noise is semantic-independent and the perturbed representations no longer represent any valid word. Moreover, adversarial perturbation requires gradient computation for each example, making it inflexible to extend to new examples.

Hard positive examples have been proven to be effective in improving model robustness~\cite{SchroffKP:15, KhoslaTWSTIMLK:20}. Thus, we hope to develop a data augmentation method that can produce hard positive examples without losing interpretability and extensibility. Based on this objective, we propose to generate hard data augmentations by weighted mixing the embedding of strong positive words with unknown-word embedding. Our motivation is to dilute the strong positive words, making the texts become more neutral and turn into hard positive examples. For example, the sentence ``{\em It's a good movie that acts exquisitely}'' expresses the {\em positive} sentiment supported by strong positive words {\em good} and {\em exquisitely}. Diluting their expressiveness makes the sentence become harder because the sentence semantics becomes less positive. Our Word dilution method is a practical realization since it is easy to implement and well interpretable. The remaining challenge is how to acquire the dilution weight assigned to each word.

To not struggle to estimate the dilution weights manually or heuristically, motivated by Generative Adversarial Networks~\cite{GoodfellowPMXWOCB:14}, we automatically learn the dilution weights and train the classifier by the constrained adversarial optimization process. Specifically, we introduce neural networks (i.e., dilution networks) to produce the dilution weights. At the inner-loop step, we fix the classifier and learn the dilution weights by maximizing the loss. At the outer-loop step, we fix the dilution weights and train the classifier by minimizing the loss with augmented inputs. We also use separate dilution networks for different classes to guide the dilution-weight learning process with the label information. As the dilution networks are learned independent of the classifier, they can be extended to compute dilution weights for new examples without further training.

To summarize, our work makes the following contributions. (1) We propose AWD data augmentation that generates hard positive examples by diluting strong positive words with unknown-word embedding. (2) We adversarially learn the dilution weights by the constrained min-max optimization with the guidance of the labels. (3) We empirically show that AWD outperforms the state-of-the-art data augmentations and demonstrates its interpretability and extensibility. 
\section{Related Works}
Low-resource text classification aims to learn a classification model from a few labeled examples. The earlier work EDA~\cite{WeiZ:19} generates text augmentations by random replacement, insertion, swap, deletion of words and trains the model on a small fraction of the original training data. The recent work explores manipulating the text data by augmenting and weighting data examples~\cite{HuTSMX:19}. Another recent work studies text data augmentations using different pre-trained transformer models~\cite{Kumar:20}. The text smoothing model converts the text data from one-hot representations to controllable smoothed representations~\cite{GLZH:22}. Following these works, we also evaluate our text augmentation method AWD in low-resource text classification. 

Text data augmentation techniques can be divided into contextual augmentation approaches and representational augmentation approaches. Besides EDA, contextual augmentation approaches also include works that substitute words using synonyms~\cite{KolomiyetsBM:11, WangY:15, MiaoLWT:20}, randomly delete, insert, replace and swap words~\cite{IyyerMBD:15, XieWLLNJN:17, ArtetxeLAC:18}, replace or re-combine fragments of the sentences~\cite{JiaL:16, Andreas:20}, paraphrase the original text~\cite{EdunovOAG:18, ChenTWG:19, KumarBBT:19, Thomas:21}, replace or generate words using language models~\cite{FadaeeBM:17a, Kobayashi:18, TavorCGKK:20, YangMFSBWBCD:20}. 

The representational augmentation approaches generate augmented inputs by interpolation or perturbation of the representations, i.e., word embedding or text representations. One popular interpolation-based representational augmentation method is Mixup~\cite{ZhangCDL:18} which generates augmented examples by linear interpolations of the pair of data points and labels. This method has recently been intensively used in NLP tasks~\cite{GuoKR:20, ZhangYZ:20}. For example, some works interpolate the original data with word-level augmented data~\cite{MiaoLWT:20} or adversarial data~\cite{ChengJME:20}, or interpolate the hidden representations of the original texts and augmented texts~\cite{ChenYY:20}. Adversarial perturbation is often used to generate perturbation-based text augmentations, e.g., applying adversarial perturbations to the word embeddings~\cite{MiyatoDG:17, ZhuCGSGL:20, ChenRLL:20, ChenJ:21} or sentence representations~\cite{HsuTG:18, WuWQ0:19}.

Improving the interpretability of adversarial perturbation methods is challenging. Recent works in this direction guide word modification by adversarial training instead of directly applying perturbation to the representations~\cite{RenDHC:19, ChengJM:19, GargR:20}. Although these methods can produce interpretable examples, the generated data augmentations sacrifice some adversarial nature by replacing the optimal adversarial representations with approximated word embeddings. They also need adversarial modification on each example, making them inflexible to extend to new examples. 
\section{Problem Formulation}
Text classification in the low-resource regime adopts supervised learning but is restricted by the available number of training examples. This task aims to efficiently learn a classification model given only a few labeled texts.

Formally, in low-resource text classification, let $\mathcal{Y}$ be the set of all classes of interest. For each class $y \in \mathcal{Y}$, we are only given a small number of $k$ labeled texts for training. Let $\mathcal{D} = \{ (x_1, y_1), (x_2, y_2), \cdots, (x_{k\times |\mathcal{Y}|}, y_{k\times |\mathcal{Y}|}) \}$ be the training set which consists of $k \times |\mathcal{Y}|$ labeled texts. Each $(x_i, y_i)$ pair in $\mathcal{D}$ denotes a text $x_i$ and its label $y_i$. Given the training set $\mathcal{D}$, the objective is to obtain a sufficiently robust model under the limited resource.

As the main challenge in this task is to deal with the overfitting problem caused by a few training examples, the previous works focus on designing various data augmentations as supplementary data to improve the model's robustness. In this paper, we follow this direction and concentrate our topic on text data augmentations in the low-resource regime. 

\section{Methodology}
In this section, we first make a general illusion of our word dilution method and the challenges in dilution weights acquisition, then introduce our systematic solution that learns the dilution weights with neural networks and trains with the text classifier by adversarial min-max optimization. 

\subsection{General Illusion of Word Dilution}
\begin{figure}[ht]
	\begin{center}
		\begin{tabular}{c}
                \hspace{-.27cm}
		    \includegraphics[width=0.47\textwidth]{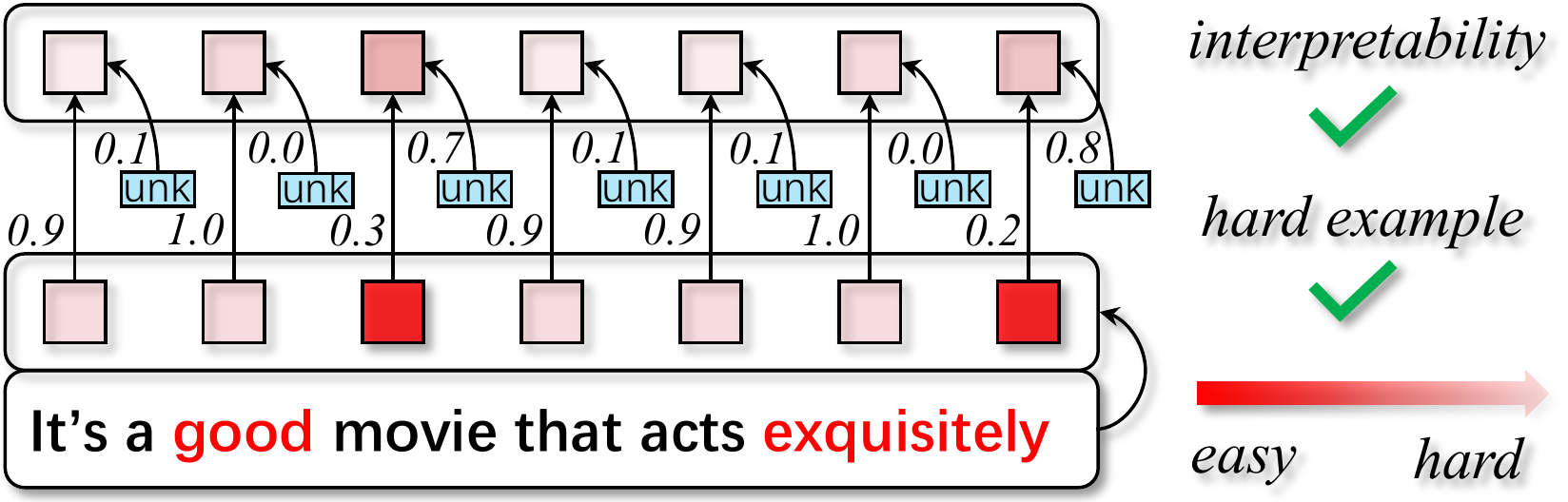}
		\end{tabular}
	\end{center}
	\caption{General illusion of word dilution. The squares represent word embeddings. The deeper color manifests that the word has more positive polarity. The weights beside the arrows represent weights of input words or the unknown word.}
	\label{fig:dilution}
\end{figure}

To generate hard positive examples as text data augmentations while keeping their semantic interpretability, we need to make word modification dependent on their semantics and simultaneously let the modified text hard to be recognized as positive by the text classifier. Considering that the polarity of a text is significantly determined by some emblematic words. As shown from the example in Figure \ref{fig:dilution}, the emblematic words ``{\em good}'' and ``{\em exquisitely}'' indicate that the sentence ``{\em It's a good movie that acts exquisitely}'' expresses the positive sentiment. We call such emblematic words {\em strong positive words}.
When we weaken the expressiveness of these strong positive words, the semantics of the text becomes more neutral and harder to be recognized by the classifier. This nature of the text data motivates us to design a new text data augmentation strategy {\em word dilution} shown in Figure \ref{fig:dilution} that dilutes the expressiveness of strong positive words in the original sentence by weighted mixing their embeddings with unknown word embedding, allowing us to generate hard positive examples without losing interpretability.

\subsubsection{Word Dilution}
For an input text example $x_i \in \mathcal{D}$, let the sequence $\{ w_{i1}, w_{i2}, \cdots, w_{in_i}\}$ be the set of all words that consist of the text, where $n_i$ denotes the total number of words in $x_i$. For each word $w_{ij}$ in $x_i$, we can obtain its word embedding $\mathbf{w}_{ij} \in \mathbb{R}^{d}$ by an embedding function $\emb$. We define the embedding function and obtain $\mathbf{w}_{ij}$ as follows
\begin{equation}
\begin{split}
\mathbf{w}_{ij} = e(w_{ij}; \mathbf{E}),
\end{split}
\label{eq:emb}
\end{equation}
where $\mathbf{E} \in \mathbb{R}^{|\mathcal{V}| \times d} $ is the parameters of the embedding function $e$, i.e., the embedding matrix, which is learnable during training. And $|\mathcal{V}|$ denotes the vocabulary size.

In a neural classification model, the word-embedding sequence $\{\mathbf{w}_{i1}, \mathbf{w}_{i2}, \cdots, \mathbf{w}_{in_i}\}$ of the input text is sent to the classifier and makes predictions. Word dilution intends to create hard positive examples challenging the classifier by weighted mixing the input word embeddings with unknown-word embedding. Specifically, we use the mark $\unk$ to denote the unknown word, whose embedding can also be obtained with the function $\emb$. We assign a dilution weight $\alpha_{ij} \in [0, 1]$ for each word $w_{ij}$ according to its relevance to the text polarity and compute the diluted word $\tilde{\mathbf{w}}_{ij}$ by
\begin{equation}
\begin{split}
\tilde{\mathbf{w}}_{ij} = (1-\alpha_{ij}) e(w_{ij}; \mathbf{E}) + \alpha_{ij} e(\unk; \mathbf{E})
\end{split}
\label{eq:dilution}
\end{equation}
After getting all diluted word embeddings $\tilde{\mathbf{w}}_{ij}$, we can treat the sequence $\{ \tilde{\mathbf{w}}_{i1}, \tilde{\mathbf{w}}_{i2}, \cdots, \tilde{\mathbf{w}}_{in_i}\}$ (simplified as $\{\tilde{\mathbf{w}}_{ij}\}$) as the data augmentation of the original input word-embedding sequence to further train the text classification model.

\subsubsection{Challenges in Dilution-Weight Acquisition}
The acquisition of dilution weights suffers two challenges. (1) The dilution weights can be obtained from experience or heuristic methods, e.g., manually selecting the strong positive words for each text or heuristically finding them according to the semantic relevance of the word embeddings. However, these methods are usually laborious and inflexible to extend to new examples. (2) An appropriate $\alpha_{ij}$ is essential to guarantee the effect of word dilution, but it is difficult to decide. We expect the high $\alpha_{ij}$ for a word closely related to the polarity of the text, i.e., the strong positive word; so that the expressiveness of these words is diluted significantly according to Equation (\ref{eq:dilution}). However, this does not mean that assigning a highest $\alpha_{ij}$ for each strong positive word is the best because an extremely high $\alpha_{ij}$ may completely obliterate the semantics of the strong positive word and make the input text meaningless or even become a false positive example. 
Such meaningless data augmentations will no longer be considered as positive examples and may harm the model training. To tackle the challenges, our solution is to adversarially learn the dilution weights with neural networks through a constrained min-max optimization process.
\begin{figure*}[ht]
	\begin{center}
		\begin{tabular}{c}
                \hspace{-.27cm}
		    \includegraphics[width=\textwidth]{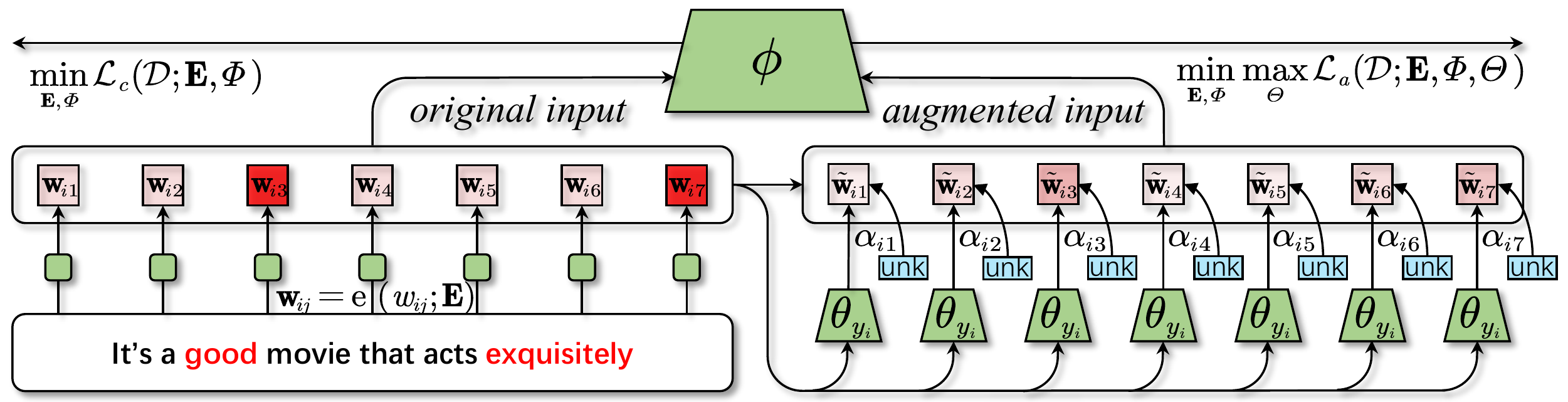}
		\end{tabular}
	\end{center}
	\caption{Adversarial word dilution. The $\phi$ and $\theta_{y_i}$ blocks respectively denote the text classifier and dilution networks.}
	\label{fig:structure}
\end{figure*}
\subsection{Systematic Solution: Adversarial Word Dilution} 
The optimization of our adversarial word dilution (AWD) is shown in Figure \ref{fig:structure}, which consists of training the classifier with original inputs and adversarially learning the dilution weights and optimizing the classifier with augmented inputs.

\subsubsection{Text Classifier}
The pre-trained language models, such as BERT~\cite{DevlinCLT:19}, have recently been employed in text classification and achieved promising results. Following previous works~\cite{HuTSMX:19, Kumar:20, GLZH:22} in low-resource text classification, we also choose BERT as the text classifier.
Specifically, BERT takes the sequence of word embeddings $\{\mathbf{w}_{i1}, \mathbf{w}_{i2}, \cdots, \mathbf{w}_{in_i}\}$ (simplified as $\{\mathbf{w}_{ij}\}$) as inputs and outputs a score corresponding to each class $y$, which we denote as $s(\{\mathbf{w}_{ij}\}, \phi_y)$, where $\phi_y$ denotes the parameters corresponding to $y$. The class $y$ with the highest score will be predicted as the label of a given text. With training set $\mathcal{D}$, we can optimize BERT by the following cross-entropy loss
\begin{equation}
\begin{split}
\!\!\mathcal{L}_{c}(\mathcal{D}; \mathbf{E}, \varPhi) &= \;\; \sum\limits_{i=1}^{k \times |\mathcal{Y}|} \!\!\ell_c(x_i, y_i; \mathbf{E}, \varPhi) \\
&= -\!\!\sum\limits_{i=1}^{k \times |\mathcal{Y}|}\!\!\log \frac{\exp(s(\{\mathbf{w}_{ij}\}, \phi_{y_i}))}{\sum_{y \in \mathcal{Y}}\exp(s(\{\mathbf{w}_{ij}\}, \phi_{y}))},
\end{split}
\label{eq:cls}
\end{equation}
where $\ell(x_i, y_i)$ is the loss corresponds to the $i^{\rm th}$ example in $\mathcal{D}$ and $\varPhi$ denotes the set of parameters $\{\phi_y: y \in \mathcal{Y}\}$.

\subsubsection{Dilution Networks}
To produce text data augmentations through word dilution, we propose the label-guided dilution networks to obtain the dilution weights $\{\alpha_{i1}, \alpha_{i2}, \cdots, \alpha_{in_i}\}$ (simplified as $\{\alpha_{ij}\}$). Our solution is two-fold: on the one hand, to prevent obtaining the dilution weights by laborious experience or heuristic methods as well as enable the data augmentations to flexibly extend to new examples, we attempt to learn the dilution weights using neural networks; on the other hand, to allow the dilution weights to be learned in accordance with the semantics of different classes, we use separate neural networks for each class and train the dilution weights guided by the labels. Specifically, for each $y \in \mathcal{Y}$, we use a Multilayer Perceptron (MLP) parameterized by $\theta_{y}$ to produce the dilution weights for all words in the input text, i.e., the dilution weight of $w_{ij}$ is computed by
\begin{equation}
\begin{split}
\alpha_{ij} = \sigma(\theta_{y_i}^{\rm T}\emb(w_{ij}; \mathbf{E})+b_{y_i}),
\end{split}
\label{eq:dilnet}
\end{equation}
where $\sigma$ is the logistic function and $b_{y_i}$ is bias. For convenience, we denote all parameters $\{\theta_y, b_y: y \in \mathcal{Y}\}$ as $\varTheta$.

\subsubsection{Adversarial Optimization} We leverage the min-max optimization derived form GAN~\cite{GoodfellowPMXWOCB:14} to adversarially update the dilution weights and train the text classifier with augmented inputs. To that end, we first modify the classification loss in Equation (\ref{eq:cls}) to include the dilution weights $\{\alpha_{ij}\}$ and augmented inputs $\{\tilde{\mathbf{w}}_{ij}\}$ as follows
\begin{equation}
\begin{split}
\!\!\!\mathcal{L}_{a}(\mathcal{D}; \mathbf{E}, \!\varPhi, \!\varTheta) &= \;\;\sum\limits_{i=1}^{k \times |\mathcal{Y}|} \!\!\ell_a(x_i, y_i, \{\alpha_{ij}\}; \mathbf{E}, \varPhi, \varTheta) \\
&= -\!\!\sum\limits_{i=1}^{k \times |\mathcal{Y}|}\!\!\log \frac{\exp(s(\{\tilde{\mathbf{w}}_{ij}\}, \phi_{y_i}))}{\sum_{y \in \mathcal{Y}}\exp(s(\{\tilde{\mathbf{w}}_{ij}\}, \phi_{y}))}
\end{split}
\label{eq:adv}
\end{equation}
To learn appropriate dilution weights, we expect not to generate extreme $\{\alpha_{ij}\}$. Thus, for each $i$, we optimize the dilution networks under the following condition 
\begin{equation}
\begin{split}
\mathcal{R}_i(\{\alpha_{ij}\}; \varTheta)=\left\| \{\alpha_{ij}\} \right\|_1-\rho n_i \leqslant 0
\end{split}
\label{eq:reg}
\end{equation}
where $\left\| \{\alpha_{ij}\} \right\|_1 \!=\! \sum_{j=1}^{n_i} |\alpha_{ij}|$ and $\rho \in (0,1)$ controls the amount of dilution allowed. This constraint guarantees that no more than $\rho$ fraction of words are diluted.

Based on Equations (\ref{eq:adv}) and (\ref{eq:reg}), we have the following constrained min-max optimization problem
\begin{equation}
\begin{split}
\!\!\min_{\mathbf{E}, \varPhi} \max_{\varTheta} \mathcal{L}_{a}(\mathcal{D}; \mathbf{E}, \varPhi, \varTheta) \;\; s.t. \sum\limits_{i=1}^{k \times |\mathcal{Y}|}\mathcal{R}_i(\{\alpha_{ij}\}; \varTheta) \leqslant 0
\end{split}
\label{eq:consopt}
\end{equation}
Note that the inner-loop constrained maximization problem can be converted to an unconstrained
problem via Penalty Function method, then the optimization problem turns to
\begin{equation}
\begin{split}
\!\!\!\min_{\mathbf{E}, \varPhi} \max_{\varTheta} [\mathcal{L}_{a}(\mathcal{D}; \!\mathbf{E}, \!\varPhi, \!\varTheta)\!-\!\lambda\!\!\!\sum\limits_{i=1}^{k \times |\mathcal{Y}|}\!\!\!\max(\mathcal{R}_i(\{\alpha_{ij}\}; \varTheta), 0)],\!\!
\end{split}
\label{eq:unconsopt}
\end{equation}
where $\lambda \geqslant 0$ is the weight of the constraint item.

It is easy to control the dilution-weight range by modifying $\rho$. However, as $\left\| \{\alpha_{ij}\} \right\|_1 \leqslant \rho n_i$ is a strict constraint, it may force the model to produce undesirable lower dilution weights for all words. To encourage generating some high dilution weights, we loose the strict constraint as

\begin{equation}
\begin{split}
\!\!\!\min_{\mathbf{E}, \varPhi} \max_{\varTheta} [\mathcal{L}_{a}(\mathcal{D}; \!\mathbf{E}, \!\varPhi, \!\varTheta)\!-\!\gamma\!\!\sum\limits_{i=1}^{k \times |\mathcal{Y}|}\!\!\left\| \{\alpha_{ij}\} \right\|_1\frac{1}{n_i}].\!\!
\end{split}
\label{eq:looseunconsopt}
\end{equation}
\vspace{-.05cm}

\begin{algorithm}[ht]
\caption{Training the Classification Model with AWD}
\hspace*{0.02in} {\bf Input:}
training set $\mathcal{D}$ and parameters $\mathbf{E}$, $\varPhi$, $\varTheta$ \\
\hspace*{0.02in} {\bf Output:}
the optimized dilution networks and classifier
\begin{algorithmic}[1]
\State initialize $\mathbf{E}$, $\varPhi$ with BERT, randomly initialize $\varTheta$;
\Repeat
    \For{each $(x_i, y_i)$ in $\mathcal{D}$}
        \State obtain $\{\mathbf{w}_{ij}\}$ through $e(w_{ij}; \mathbf{E})$;
    \EndFor
    \State input all $\{\mathbf{w}_{ij}\}$, update $\mathbf{E}$, $\varPhi$ by $\min\mathcal{L}_{c}(\mathcal{D}; \mathbf{E}, \varPhi)$;
    \State fix $\mathbf{E}$, $\varPhi$, update $\varTheta$ by $\max\mathcal{L}_{a}(\mathcal{D}; \mathbf{E}, \varPhi, \varTheta)$;
    \State compute all $\{\alpha_{ij}\}$, generate all $\{\tilde{\mathbf{w}}_{ij}\}$ by eq. (\ref{eq:dilution});
    \State fix $\varTheta$, update $\mathbf{E}$, $\varPhi$ by $\min\mathcal{L}_{a}(\mathcal{D}; \mathbf{E}, \varPhi, \varTheta)$;
\Until{convergence}
\end{algorithmic}
\label{alg:1}
\end{algorithm}
\subsection{Training Text Classification Model with AWD}
We train the text classification model with adversarial word dilution by iteratively executing three optimization steps shown in Algorithm \ref{alg:1} (to describe the process easier, we omit the constraint items in (\ref{eq:unconsopt}) and (\ref{eq:looseunconsopt})): (1) input the original data and minimize $\mathcal{L}_{c}(\mathcal{D}; \mathbf{E}, \varPhi)$; (2) fix $\mathbf{E}$, $\varPhi$ and update $\varTheta$ by maximizing $\mathcal{L}_{a}(\mathcal{D}; \mathbf{E}, \varPhi, \varTheta)$, compute dilution weights and generate augmented data; (3) fix $\varTheta$, input the augmented data and update $\mathbf{E}$, $\varPhi$ by minimizing $\mathcal{L}_{a}(\mathcal{D}; \mathbf{E}, \varPhi, \varTheta)$. We perform one SGD updating at each optimization step.

For convenience, we name the model optimized with the strict constraint term in (\ref{eq:unconsopt}) as {\bf AWD(strict)} and loosed constraint item in (\ref{eq:looseunconsopt}) as {\bf AWD} or {\bf AWD(loose)}, respectively.

\section{Experiment} 
\subsection{Datasets and Experiment Setting}
Following the previous works in low-resource text calssification~\cite{WeiZ:19, HuTSMX:19, Kumar:20, GLZH:22}, we evaluate our data augmentations on three benchmarks: SST, TREC, and SNIPS.
\subsubsection{SST-2}
Stanford Sentiment Treebank (SST)~\cite{SocherPWCMNP:13} is a sentiment classification dataset. The text data are extracted from movie reviews of the rottentomatoes.com website and is originally collected by \cite{PangL:05}. The texts involve $2$ classes, i.e., positive and negative.
\subsubsection{TREC}
TREC~\cite{LiR:02} is a fine-grained question classification dataset. The text data is collected from USC, TREC 8, TREC 9 and TREC 10 questions. The $500$ questions from TREC 10 are used for test. This dataset contains $6$ question types (including person, location, etc.).
\subsubsection{SNIPS}
SNIPS~\cite{Coucke:18} is an English dataset for natural language understanding, which is widely used in intent classification. The text data is collected from crowd-sourced queries. This dataset contains $7$ user intents from different domains (such as movie, book, etc.).
\subsubsection{Experiment Setting}
We use the datasets provided by \cite{GLZH:22}. To simulate low-resource text classification, we randomly select $k\!=\!10, 20, 50$ examples for each class as the training sets. The training set with $\!k=\!10$, the validation set and the test set are the same in \cite{GLZH:22}. The data statistics are shown in Table \ref{tab:statics}.
\begin{table}[ht]
	\centering
    \resizebox{\linewidth}{!}{
	\begin{tabular}{cccccc}
		\toprule
		Dataset & \#train & \#low-res & \#val & \#test & \#class\\
		\midrule
		SST & 6,228 & 10/20/50 & 20 & 1821 & 2\\
		TREC & 5,406 & 10/20/50 & 60 & 500 & 6\\
		SNIPS & 13,084 & 10/20/50 & 70 & 700 & 7\\
		\bottomrule
	\end{tabular}
        }
	\caption{The statistics of the datasets. 
 }
	\label{tab:statics}
\end{table}
\subsection{Baseline Models and Implementation}
\subsubsection{Baseline Models} 
We compare our AWD with the following baseline models: the pure {\bf BERT}~\cite{DevlinCLT:19} without data augmentations; contextual augmentation by word modification or paraphrasing, including {\bf EDA}~\cite{WeiZ:19} and {\bf BT}(Back Translation)~\cite{Shleifer:19}; contextual augmentation using pre-trained laguage models by word modification, including {\bf CBERT}~\cite{WuLZHH:19}, {\bf BERTexpand}, {\bf BERTprepend}~\cite{Kumar:20}, and by generation, including {\bf GPT2}; representational data augmentation including {\bf Mixup}~\cite{ZhangCDL:18}, {\bf Textsmooth}~\cite{GLZH:22}, {\bf ADV}(adversarial data augmentation)~\cite{MiyatoDG:17}.
\subsubsection{Implementation} 
We implement our AWD model using Pytorch deep learning framework\footnote{Code: https://github.com/BDBC-KG-NLP/AAAI2023\_AWD}. The BERT-uncased Base model is used as the text classifier. The dimension of the word embedding $d$ is set to $768$. The dilution network for each class is implemented using an MLP followed by a sigmoid activation function. We train our model using Adam optimizer with default configurations. The learning rate is set to $5\times 10^{-4}$. We train AWD and each baseline model for $30$ epochs. We repeat all experiments $15$ times and report their mean accuracy. We tune the hyper-parameters by grid search on the validation set. The hyper-parameter for training AWD(strict) is $\lambda\!\!=\!\!1$ and $\rho\!\!=\!\!0.3, 0.5, 0.3$ for respect $k\!\!=\!\!10, 20, 50$. When training AWD(strict), we perform $5$ SGD updates in a dilution-network optimization step with a learning rate of $0.01$. The hyper-parameter for training AWD(loose) is $\gamma=0.005$. All experiments are conducted on an NVIDIA Tesla P100 GPU with $16$GB memory. 
\begin{table*}[ht]
	\centering
	\setlength{\tabcolsep}{3pt}
    \resizebox{\linewidth}{!}{
	\begin{tabular}{ccccccccccccc}
		\toprule
		\multirow{2}*{\textbf{Method}}&\multicolumn{3}{c}{\textbf{SST-2}}&\multicolumn{3}{c}{\textbf{TREC}}&\multicolumn{3}{c}{\textbf{SNIPS}}&\multicolumn{3}{c}{\textbf{Average}}\\
		\cmidrule(lr){2-4} \cmidrule(lr){5-7} \cmidrule(lr){8-10}\cmidrule(lr){11-13} &
		$k\!\!=\!\!10$& $k\!\!=\!\!20$ & $k\!\!=\!\!50$ & $k\!\!=\!\!10$& $k\!\!=\!\!20$ & $k\!\!=\!\!50$ & $k\!\!=\!\!10$& $k\!\!=\!\!20$ & $k\!\!=\!\!50$ & $k=10$& $k=20$ & $k=50$\\
		\midrule
		BERT & 62.2(7.1) & 71.9(4.8) & 79.7(4.3) & 72.1(10.5) & 82.9(4.9) & 88.4(3.4) & 90.6(1.3) & 92.9(1.4) & 94.7(0.9) & 74.9(6.3) & 82.6(3.7) & 87.6(2.9)\\
		\midrule
		EDA & 62.7(8.5) & 71.5(6.1) & 80.3(3.1) & 75.0(7.5) & 80.8(4.6) & 86.6(3.9) & 90.2(2.0) & 93.1(1.2) & 94.2(1.3) & 75.9(6.0) & 81.8(4.0) & 87.1(2.8)\\
		BT & 64.0(7.8) & 70.1(6.8) & 80.8(3.8) & 73.7(7.8) & 82.1(5.9) & 88.5(2.3) & 91.0(1.5) & 93.2(0.9) & 94.6(0.8) & 76.3(5.7) & 81.8(4.6) & 88.0(2.3)\\
		CBERT & 61.4(7.9) & 72.6(6.6) & 80.2(3.1) & 73.2(7.9) & 82.7(5.0) & 88.7(2.7) & 90.1(2.3) & 93.2(1.0) & 94.3(1.2) & 74.9(6.1) & 82.9(4.2) & 87.7(2.3)\\
		BERTexpand & 62.4(8.3) & 71.1(5.2) & 80.3(4.5) & 75.3(6.4) & 83.3(4.6) & 86.7(4.0) & 90.5(2.1) & 93.1(1.3) & 94.8(1.0) & 76.0(5.6) & 82.5(3.7) & 87.3(3.2) \\
		BERTprepend & 63.8(7.6) & 70.9(5.4) & 78.8(6.2) & 73.1(7.5) & 81.4(4.4) & 86.0(3.7) & 90.4(1.7) & 93.4(1.1) & 94.9(1.1) & 75.7(5.6) & 81.9(3.6) & 86.6(3.6) \\
		GPT2 & 63.8(5.4) & 69.7(5.9) & 75.7(5.0) & 74.2(7.2) & 80.2(6.9) & 84.8(4.1) & 90.3(1.4) & 93.2(0.6) & 93.5(1.3) & 76.1(4.7) & 81.0(4.5) & 84.7(3.5) \\
            \midrule
		Mixup & 64.4(5.8) & 72.6(4.0) & 80.3(6.8) & 73.6(5.7) & 81.3(6.0) & 87.8(4.2) & {\bf 91.4}(1.1) & {\bf 93.9}(1.2) & 94.8(0.8) & 76.5(4.2) & 82.6(3.7) & 87.6(3.9) \\
		Textsmooth & 63.1(7.6) & 72.5(5.6) & 82.5(2.8) & 75.3(6.7) & 81.3(5.8) & 88.1(3.9) & 90.5(1.2) & 93.7(1.0) & 94.8(1.0) & 76.3(5.2) & 82.5(4.2) & 88.5(2.6) \\
		ADV & 64.0(7.7) & 72.4(8.1) & 81.3(3.9) & 74.9(9.6) & 82.6(6.4) & 88.0(4.7) & 90.5(2.4) & 93.5(0.9) & 94.8(0.7) & 76.5(6.6) & 82.8(5.1) & 88.0(3.1) \\
		{\bf AWD(strict)} & {\bf 65.4}(6.8) & \underline{72.9}(5.8) & \underline{82.7}(3.0) & \underline{76.7}(7.3) & {\bf 83.7}(4.4) & \underline{89.1}(3.4) & \underline{91.2}(1.8) & \underline{93.8}(1.0) & \underline{95.0}(1.0) & \underline{77.7}(5.3) & \underline{83.5}(3.7) & \underline{88.9}(2.5) \\
		{\bf AWD(loose)} & \underline{65.2}(7.8) & {\bf 73.9}(5.4) & {\bf 82.7}(5.2) & {\bf 77.1}(7.1) & \underline{83.4}(4.2) & {\bf 90.0}(3.1) & 91.0(1.9) & 93.7(1.2) & {\bf 95.3}(0.7) & {\bf 77.8}(5.6) & {\bf 83.7}(3.6) & {\bf 89.3}(3.0) \\
		\bottomrule
	\end{tabular}
	}
	\caption{The evaluation results on SST-2, TREC and SNIPS. The bold and underline indicate the best and second-best results.}
	\label{tab:main}
\end{table*}
\subsection{Low-Resource Text Classification Results}
\subsubsection{Main Results} 
The low-resource text classification results of AWD and baseline models are shown in Table \ref{tab:main}. The results show that most contextual data augmentation methods and representational data augmentation methods improve upon the BERT baselines. The representational data augmentations are relatively better than the contextual data augmentation methods. These observations manifest the effectiveness of representational data augmentations. Compared to existing representational data augmentation methods Mixup, Textsmooth and ADV, our AWD(strict) and AWD(loose) achieve better results in all settings on SST-2, TREC, Average and in the $k\!=\!50$ setting on SNIPS. These results demonstrate that our methods AWD(strict) and AWD(loose) significantly improve generalization in low-resource text classification and build themselves as the new state-of-the-art methods of text data augmentation. 
\begin{figure}[ht]
	\begin{center}
		\begin{tabular}{cc}
		\hspace{-.25cm}
			\resizebox{0.5\linewidth}{!}{
		    \includegraphics{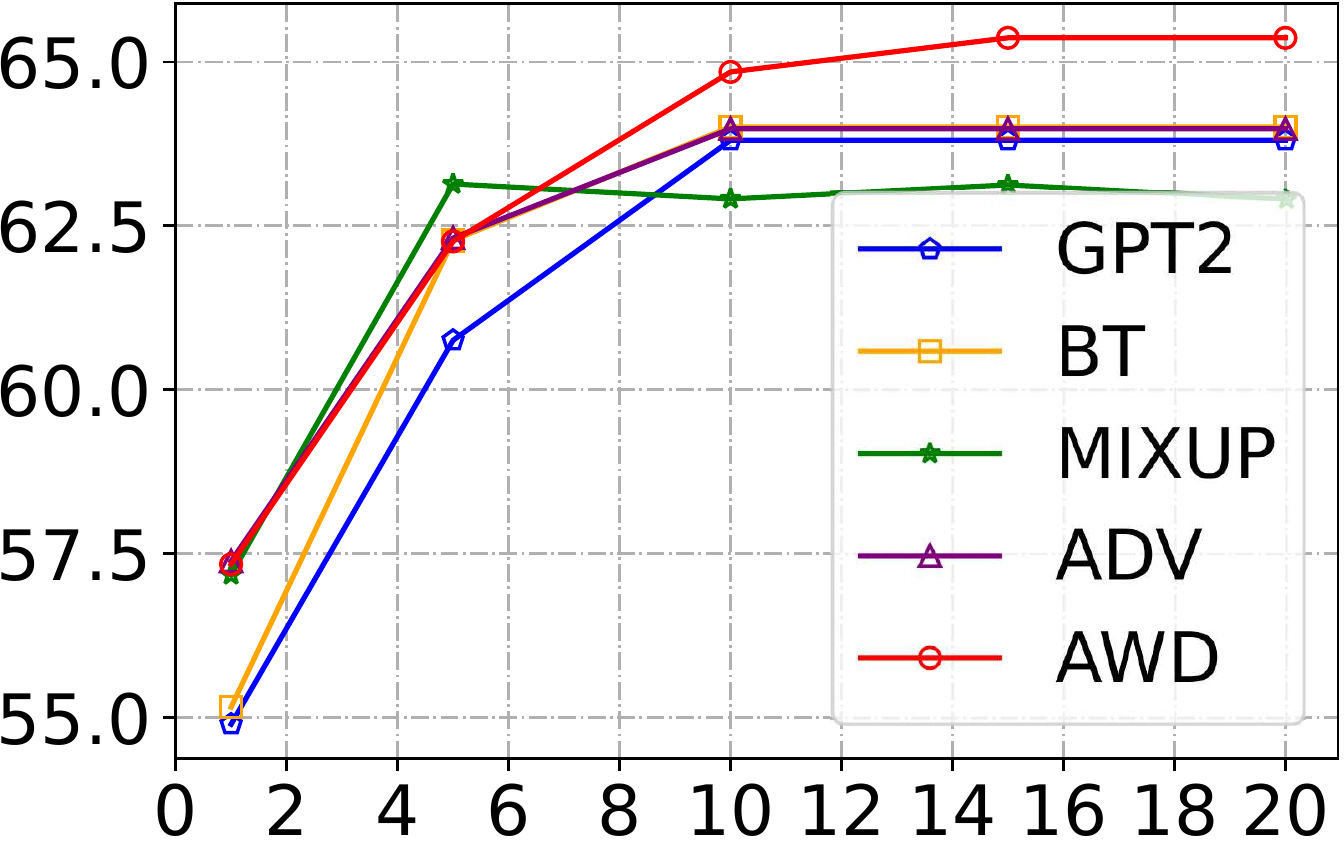}
			}&
		\hspace{-.45cm}
			\resizebox{0.472\linewidth}{!}{
			\includegraphics{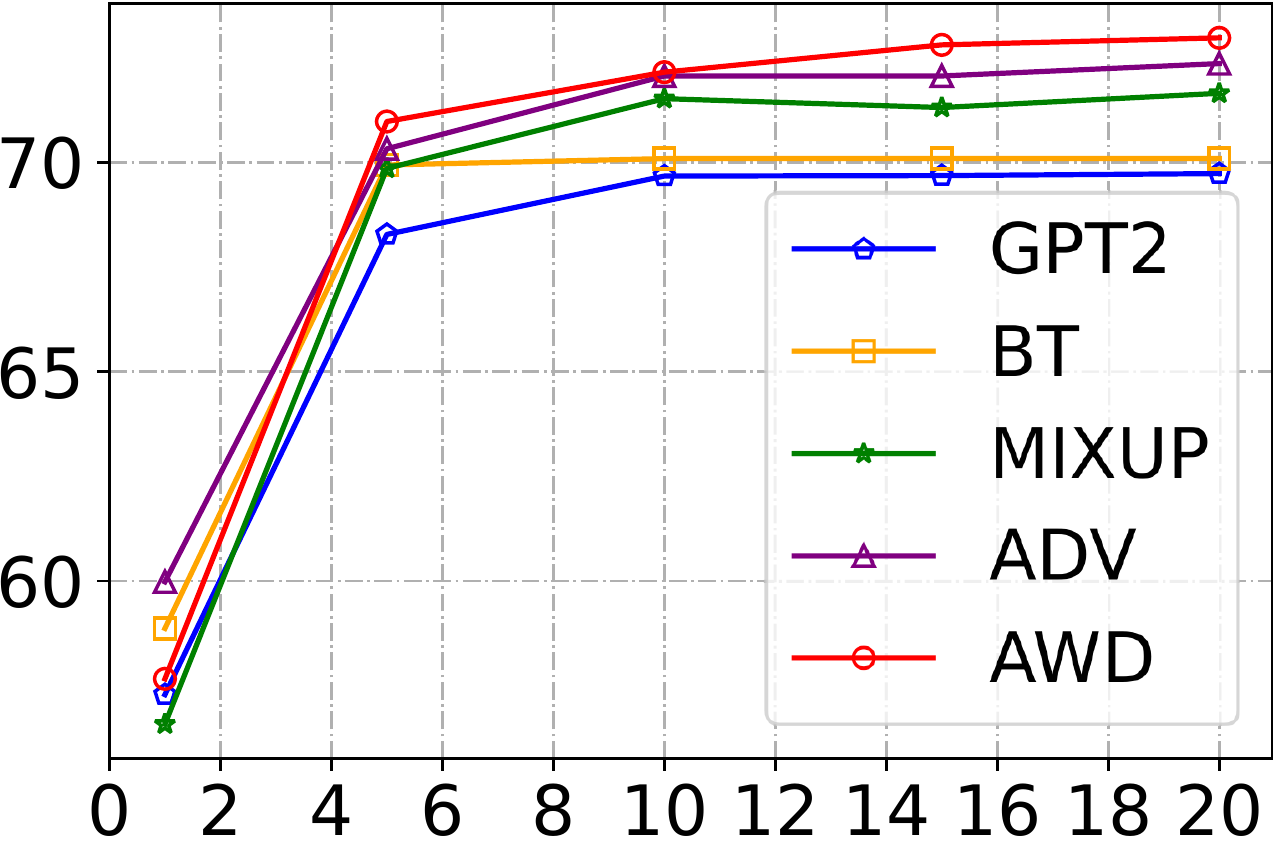}
			}\\
		\hspace{-.05cm}
		(a) $k$=10 & 
		\hspace{-.3cm}
		(b) $k$=20 
		\end{tabular}
	\end{center}
	\caption{Results during the training. The horizontal and vertical axes respectively denote the epochs and accuracy.}
	\label{fig:eff}
\end{figure}
\subsubsection{The Effectiveness of Different Data Augmentations} 
To investigate the effectiveness of training models using different data augmentations,
we compare the evaluation accuracy after different epochs during the training process on SST-2 in Figure \ref{fig:eff}. As shown in the figure, our AWD generally achieves the best accuracy as the training progresses in both $k\!=\!10$ and $k\!=\!20$ settings, which manifests the effectiveness of AWD data augmentations. Another observation is that most data augmentations except AWD converge near $10$ epochs and their accuracy stops increasing. Our AWD continues to increase the model performance after $10$ epochs. This observation demonstrates that AWD can produce more effective text data augmentations that boost the model training. It also implicitly suggests that AWD may produce hard positive examples that benefit the training of the later stage. 

\subsection{More Detailed Analysis on AWD}
\subsubsection{The Hardness of Different Data Augmentations} 
We train a BERT model using the whole training set excluding the low-resource data as a discriminator to measure the hardness of given data augmentations, i.e., a higher error rate on this discriminator implies that the examples are harder. To appropriately evaluate all methods, we compute the harmonic mean (HM), the integrated performance of classification accuracy (Acc) and error rate on the discriminator (Err). Harder augmentations and better accuracy result in High HM. High Err but low Acc reflect that a model may produce undesired hard negative augmentations. 
\begin{figure}[ht]
	\begin{center}
		\begin{tabular}{cc}
		\hspace{-.29cm}
			\resizebox{0.5\linewidth}{!}{
		    \includegraphics{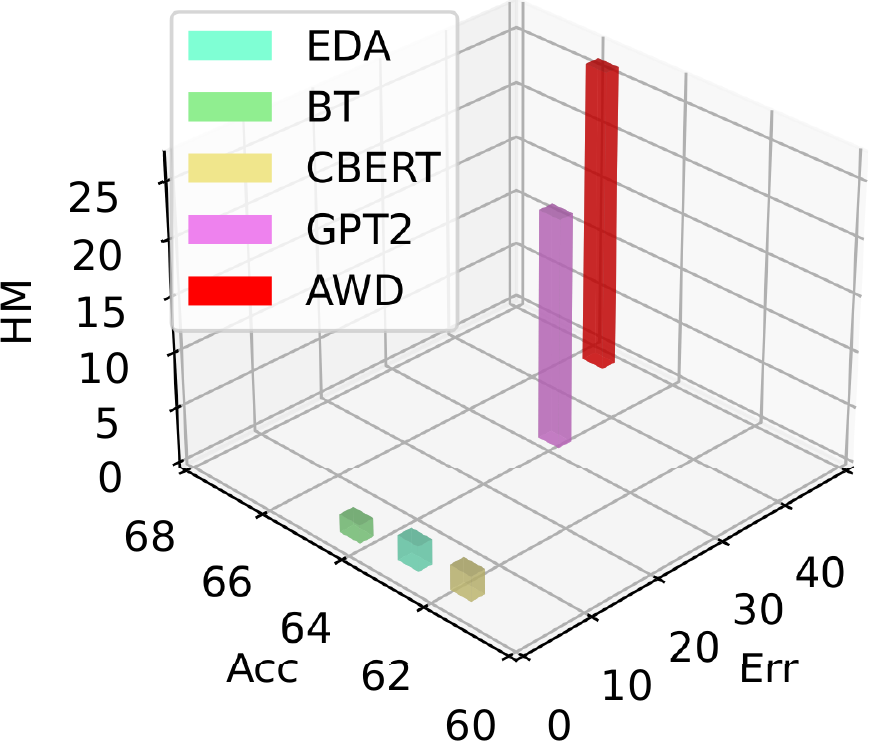}
			}&
		\hspace{-.59cm}
			\resizebox{0.5\linewidth}{!}{
			\includegraphics{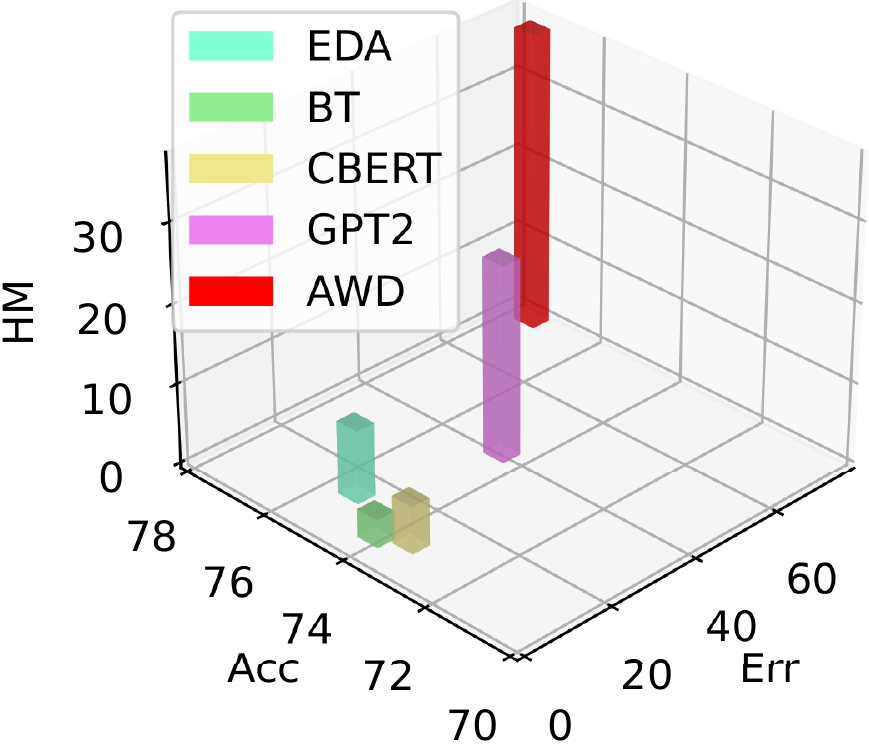}
			}\\
		\hspace{-.09cm}
		(a) SST-2 & 
		\hspace{-.39cm}
		(b) TREC
		\end{tabular}
	\end{center}
	\caption{The hardness analysis of different text data augmentation methods on the SST-2 and TREC datasets.}
	\label{fig:hard}
\end{figure}
We compare different text data augmentations with $k\!=\!10$ on SST-2 and TREC in Figure \ref{fig:hard}, illustrating that harder data augmentations may increase accuracy, but there are exceptions, e.g., GPT2 produces relatively harder data augmentations but has lower accuracy. It may be because it produces hard but negative examples that may harm the model training. Among the compared models, our AWD achieve the highest HM because of high Acc and Err, demonstrating that AWD effectively produces hard positive examples.

\subsubsection{Extending AWD to New Examples} 
To study the extensibility of AWD to new examples, we first pre-train the AWD(strict) and AWD(loose) models on $k\!=\!10,20,50$ training sets, then fix their parameters and use their dilution networks to generate data augmentations for new examples in new training sets with respect $k\!=\!10, 20, 50$, next train the BERT model on each new training sets together with the respectively generated data augmentations. Note that in this setup, the dilution weights are not learned from the training data but extended from the data used for AWD pre-training. The evaluation results of BERT trained with original data and extended augmented data are shown in Table \ref{tab:new}, which presents that the data augmentations for new examples generated by our AWD models effectively improve the original BERT model. This suggests that AWD can extend well to new examples without further training. 
\begin{table*}[ht]
	\centering
	\setlength{\tabcolsep}{3pt}
    \resizebox{\linewidth}{!}{
	\begin{tabular}{ccccccccccccc}
		\toprule
		\multirow{2}*{\textbf{Method}}&\multicolumn{3}{c}{\textbf{SST-2}}&\multicolumn{3}{c}{\textbf{TREC}}&\multicolumn{3}{c}{\textbf{SNIPS}}&\multicolumn{3}{c}{\textbf{Average}}\\
		\cmidrule(lr){2-4} \cmidrule(lr){5-7} \cmidrule(lr){8-10}\cmidrule(lr){11-13} &
		$k\!\!=\!\!10$& $k\!\!=\!\!20$ & $k\!\!=\!\!50$ & $k\!\!=\!\!10$& $k\!\!=\!\!20$ & $k\!\!=\!\!50$ & $k\!\!=\!\!10$& $k\!\!=\!\!20$ & $k\!\!=\!\!50$ & $k=10$& $k=20$ & $k=50$\\
		\midrule
		BERT & 62.2(7.1) & 71.9(4.8) & 79.7(4.3) & 72.1(10.5) & 82.9(4.9) & 88.4(3.4) & 90.6(1.3) & 92.9(1.4) & 94.7(0.9) & 74.9(6.3) & 82.6(3.7) & 87.6(2.9)\\
		{\bf BERT+strict\;} & {\bf 63.2}(7.8) & 72.7(6.2) & 81.0(3.6) & {\bf 75.2}(9.3) & {\bf 83.5}(4.5) & {\bf 89.2}(3.2) & {\bf 90.9}(2.3) & 93.3(1.1) & 94.6(0.9) & {\bf 76.4}(6.5) & 83.2(3.9) & 88.3(2.6) \\
		{\bf BERT+loose\;} & 63.2(7.5) & {\bf 74.2}(5.7) & {\bf 82.5}(2.2) & 74.8(9.7) & 83.2(4.5) & 88.9(2.2) & 90.8(1.7) & {\bf 93.4}(1.0) & {\bf 94.9}(1.1) & 76.2(6.3) & {\bf 83.6}(3.7) & {\bf 88.8}(1.8) \\
		\bottomrule
	\end{tabular}
	}
	\caption{The results of applying AWD models to new examples. BERT+strict and BERT+loose denote the BERT model trained with data augmentations for new examples generated by pre-trained AWD(strict) and AWD(loose), respectively.}
	\label{tab:new}
\end{table*}

\begin{figure}[ht]
	\begin{center}
		\begin{tabular}{cc}
		\hspace{-.25cm}
			\resizebox{0.475\linewidth}{!}{
		    \includegraphics{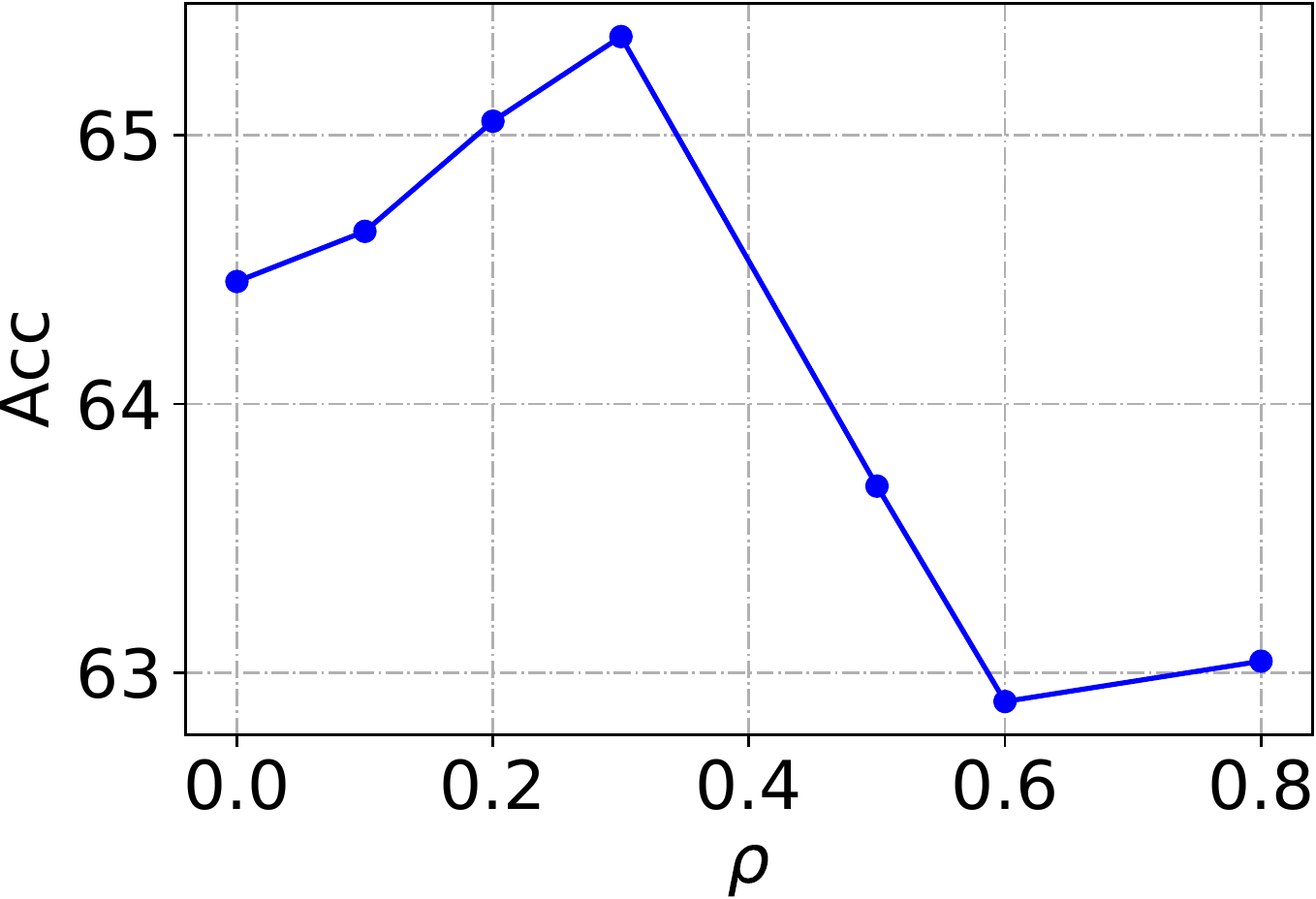}
			}&
		\hspace{-.55cm}
			\resizebox{0.5\linewidth}{!}{
			\includegraphics{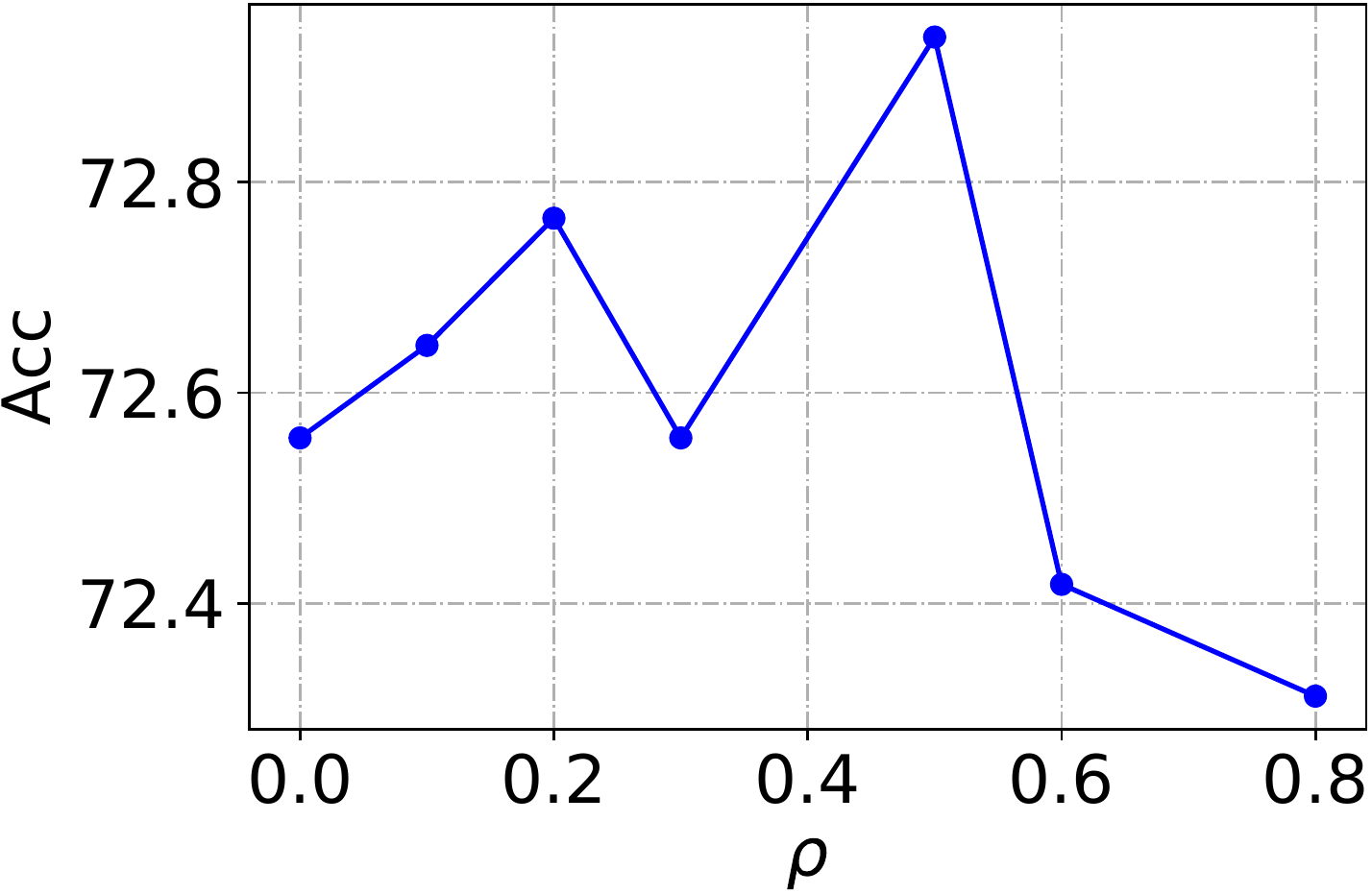}
			}\\
		\hspace{-.25cm}
		(a) $k$=10 & 
		\hspace{-.55cm}
		(b) $k$=20
		\end{tabular}
	\end{center}
	\caption{The hyper-parameter analysis of $\rho$ on SST-2.}
	\label{fig:hyper}
\end{figure}
\subsubsection{The Hyper-Parameter Analysis of AWD} 
As mentioned in the challenges, choosing appropriate dilution weights is essential to guarantee the effect of word dilution. Extreme dilution weights may lead the augmented texts to be useless or meaningless. To investigate this phenomenon, we illustrate the performance changes with various scopes of dilution weights by controlling the hyper-parameter $\rho$ in the AWD(strict) model. We report the results on SST-2 using different hyper-parameter $\rho$ in Figure \ref{fig:hyper}. As shown in the figure, In both $k\!\!=\!\!10$ and $k\!\!=\!\!20$ settings, the model performance generally increases first and then decreases as the hyper-parameter $\rho$ increase. The illustrated tendencies suggest our guess. When $\rho$ is too small, the model generates low dilution weights and constructs data augmentations with minor-difference from the original texts that may bring slight improvement. When $\rho$ is too large, the model generates high dilution weights, making the augmented texts become meaningless and may degrade the performance.

\begin{figure}[ht]
    \begin{center}
        \resizebox{\linewidth}{!}{
            \includegraphics{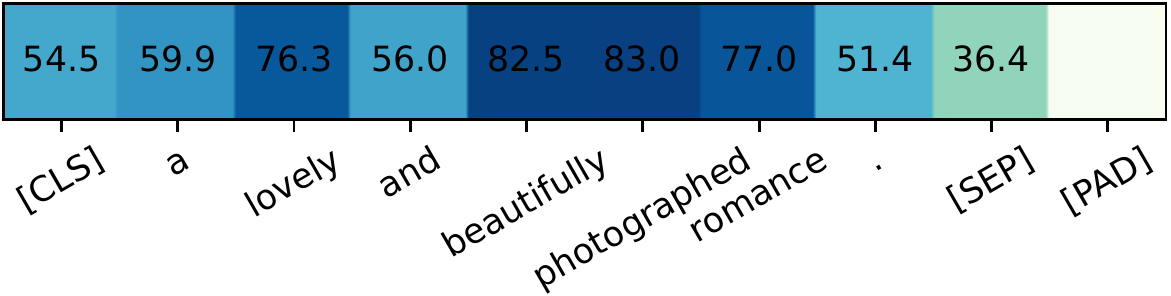}
        }
        (a) An example in SST-2 with label {\em Positive}. \\
        \resizebox{\linewidth}{!}{
            \includegraphics{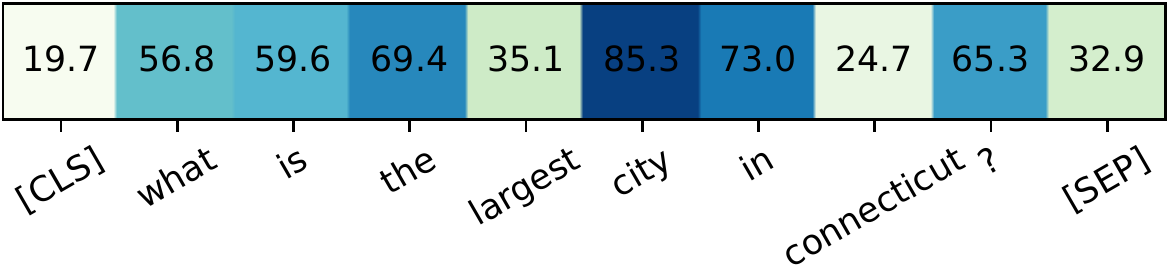}
        }
        (b) An example in TREC with label {\em Location}.\\
        \resizebox{\linewidth}{!}{
            \includegraphics{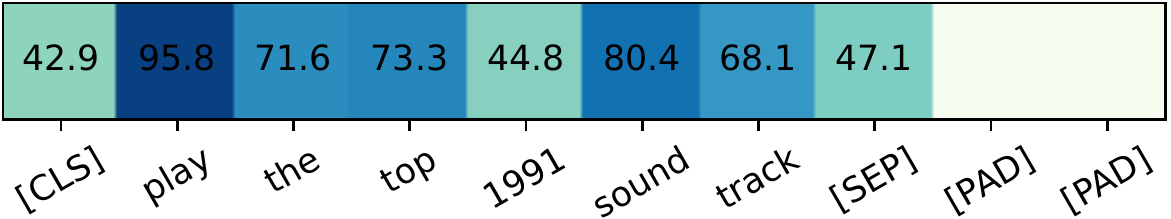}
        }
        (c) An example in SNIPS with label {\em PlayMusic}.
    \end{center}
	\caption{The selected examples from the three datasets, the values indicate the corresponding dilution weights. We pad the three sentences to the same length for a better view.}
	\label{fig:case}
\end{figure}
\subsubsection{The Interpretability of AWD} 
To demonstrate the interpretability of AWD, we make case studies on the dilution weights assigned to each word of some selected examples. We select three text examples, each from one of the SST-2, TREC, SNIPS datasets, and visualize their learned dilution weights in Figure \ref{fig:case}. In the (a) case from the SST-2 dataset, the words {\em lovely}, {\em beautifully}, {\em photographed} and {\em romance} all express positive sentiment and they are semantically related to the label {\em Positive} of the text. These words are respectively assigned with high dilution weights $0.763$, $0.825$, $0.830$ and $0.770$. This case manifests that our AWD model effectively learns the interpretable dilution weights through adversarial optimization. In the (b) case from the TREC dataset, the word {\em city} related to the label {\em Location} is assigned with a high dilution weight $0.853$. However, the word {\em connecticut} also related to {\em Location} is assigned a low dilution weight $0.247$. This may be because this word is a low-frequency knowledge word whose semantics are difficult to capture. In the (c) case from the SNIPS dataset, the words {\em play} and {\em sound} related to the label {\em PlayMusic} are assigned with high dilution weights. All special words, such as {\em [CLS]}, {\em [SEP]} are assigned with relatively low dilution weights. These cases demonstrate that the dilution weights generated by our AWD model are semantically interpretable.

\section{Conclusion}
This work proposes a new text data augmentation method, Adversarial Word Dilution (AWD), for low-resource text classification. It can generate hard positive examples as data augmentations while keeping interpretability and extensibility. AWD constructs data augmentations through word dilution by weighted mixing the embedding of strong positive words with unknown-word embedding. We adversarially train AWD with constrained min-max optimization by iteratively learning the dilution weights and training the classifier. Experiment results show that AWD outperforms the state-of-the-art text data augmentation methods and demonstrates its interpretability and extensibility. We hope AWD will extend to a broad spectrum of other text classification tasks with limited labeled data,  such as few-shot text classification or semi-supervised text classification.

\section*{Acknowledgments}
This work was supported in part by the National Key R\&D Program of China under Grant 2021ZD0110700, in part by the Fundamental Research Funds for the Central Universities, in part by the State Key Laboratory of Software Development Environment.

\bibliography{aaai23}

\end{document}